\documentclass[runningheads]{llncs_custom}
\usepackage{amssymb}
\usepackage{amsmath}
\usepackage[T1]{fontenc}
\usepackage{booktabs}
\usepackage{multirow}
\usepackage{tabularx}  
\usepackage[linktoc=all]{hyperref}
\usepackage{graphicx}
\usepackage{caption}
\usepackage{color}

\newcolumntype{Y}{>{\centering\arraybackslash}X}
\begin{document}

\title{A Unified Platform to Evaluate STDP Learning Rule and Synapse Model using Pattern Recognition in a Spiking Neural Network}

\titlerunning{Unified Evaluation of STDP \& Synapse Models in SNNs}

\author{Jaskirat Singh Maskeen\inst{1}\orcidID{0009-0001-6781-8152} \and \\
Sandip Lashkare\inst{2}\orcidID{0000-0003-2018-1681}}

\authorrunning{J. S. Maskeen and S. Lashkare}
%
\institute{Indian Institute of Technology Gandhinagar, India \\
\email{\{23110146, sandip.lashkare\}@iitgn.ac.in}}

\maketitle              
\begin{abstract}
We develop a unified platform to evaluate Ideal, Linear, and Non-linear $\text{Pr}_{0.7}\text{Ca}_{0.3}\text{MnO}_{3}$ memristor-based synapse models, each getting progressively closer to hardware realism, alongside four STDP learning rules in a two-layer SNN with LIF neurons and adaptive thresholds for five-class MNIST classification. On MNIST with small train set and large test set, our two-layer SNN with ideal, 25-state, and 12-state non-linear memristor synapses achieves 92.73 \%, 91.07 \%, and 80 \% accuracy, respectively, while converging faster and using fewer parameters than comparable ANN/CNN baselines.

\end{abstract}

\section{Introduction}

Spiking neural networks (SNNs) promise brain-inspired efficiency by co-locating memory and computation, overcoming the memory-bandwidth bottleneck inherent to Von-Neumann processors~\cite{nwad283,50211040}. Neuromorphic hardware such as memristor crossbars implements synaptic updates locally and in parallel, yielding orders of magnitude energy and speed improvements over CPUs/GPUs~\cite{50211040}. Memristors like Resistive Random-Access Memory (RRAM) have positioned them as leading on-chip synapse candidates due to high density, nonvolatility, and precise conductance control~\cite{Prezioso2015-sw}.

In this work, we target pattern recognition under tight data constraints. Unlike SNN works using surrogate-gradient backpropagation~\cite{XU201399,6797168,SHEN2022100522}, which require global error signals and centralized weight updates (complicating deployment on neuromorphic hardware), we adopt a supervised Hebbian rule derived from local Spike-Timing-Dependent-Plasticity (STDP)~\cite{Kasinski2006} for entirely local synaptic plasticity. Our custom Python SNN simulator\footnote{\url{https://github.com/jsmaskeen/nervos}} supports arbitrary finite-state synapse models and Leaky Integrate-and-Fire (LIF) neurons with adaptive thresholds for spike-frequency adaptation as observed in biology~\cite{Platkiewicz_Brette_2007}. We use the MNIST dataset~\cite{726791} to evaluate how hardware-realistic synapse models and local STDP rules affect accuracy and efficiency. While Convolutional Neural Networks (CNNs) can surpass 99\% on MNIST~\cite{726791}, they need extensive labeled data and significant computational resources for backpropagation. In contrast, SNNs leverage sparse, event-driven spikes and unsupervised STDP to learn features with far fewer samples and compute resources~\cite{fncom.2015.00099}.

Several recent studies achieve over 97\% on MNIST by using more complex STDP variants in deeper SNNs, for example, Stabilized Supervised STDP (S2-STDP) in multi-layer convolutional SNNs and reward-modulated STDP (R-STDP) in deep spiking networks~\cite{fnins.2024.1401690,8963900,ZHAO202268}.  However, those approaches rely on multi-layer architectures and do not evaluate how a minimal two-layer SNN would perform. Moreover, the analysis of the impact of hardware-realistic synapse quantization or finite-state nonlinearity on classification accuracy, along with different STDP models, is limited.

Our work bridges this gap by presenting a unified platform to evaluate the SNN performance by comparing four STDP variants: \textbf{Conventional}~\cite{Bi10464}, \textbf{Cosine-shaped}, \textbf{Sinusoidal-shaped}, \textbf{Gaussian-shaped}, paired with three synapse models: \textbf{Ideal} (theoretical upper bound for accuracy), \textbf{Linear} (with finite states), and \textbf{Non-linear} (with finite states) based on Pr$_{0.7}$Ca$_{0.3}$MnO$_3$ data~\cite{Lashkare_2020}, in a simple two-layer network. We use only local STDP (no backpropagation), which allows us to systematically quantify accuracy, convergence speed, and hardware-relevant trade-offs under tight data constraints on the MNIST dataset. We also track synaptic weight changes to identify emerging patterns. By focusing on a minimal architecture and realistic synapse behavior, our study provides a reference to the neuromorphic community for deploying on-chip SNNs where deep architectures or backpropagation are impractical.
\section{Methodology}

\subsection{Network Architecture}
We use a two-layer SNN, with no hidden layers.
Layer 1 has 784 neurons, and Layer 2 has 60 neurons (with Non-linear Synapse) or 80 neurons (with Ideal Synapse), for five-class 
classification, and 80 neurons for ten-class classification. 
Excitatory synapses connect each input neuron to every output neuron, 
using models described in subsequent sections. All neurons in the output layer are 
connected amongst themselves using lateral inhibitory synapses, and utilise
the \textit{Winner-Takes-All} rule~\cite{Gupta_2007}. So, after a training image
has been passed through the network for a fixed duration,
the output neuron that fires the most is assigned the label of the training
image. So, at the end of training, each output neuron corresponds to a particular label.
\subsection{Neuron Model}
We use a modified Leaky Integrate-and-Fire (LIF) model, which incorporates
a biologically inspired mechanism of adaptive thresholding, which prevents
rapid firing of neurons by increasing the threshold after a spike, which decays
back to the initial threshold potential with some time constant $\tau_0$~\cite{koch2004methods,Fontaine_2014}.

It is a simple resistor-capacitor circuit that models the dynamics of neuronal
membrane potential. The neuron fires a spike once the membrane potential
crosses a threshold ($V_{th}$). After each spike, the membrane potential
is reset to $V_{reset}$, while the threshold adapts dynamically based on
recent neuronal activity. This adaptive process momentarily elevates the
threshold to inhibit rapid consecutive firing, then gradually decays 
to its baseline value with a time constant $\tau_0$.\\

\noindent\textbf{Membrane Dynamics:} 
Equation~\ref{eqn:lifdynamics} describes the dynamics of our neuron's membrane~\cite{koch2004methods}.
\begin{equation}
    \label{eqn:lifdynamics}
    \frac{dV}{dt} = \frac{1}{C_m}(-g(V-E_L) + I)
\end{equation}
where $V$ is membrane potential, $C_m$ is capacitance of membrane, $g$ is
conductance, $E_L$ is resting potential and $I$ is the input current to the
neuron.\\
Values used: $C_m = 8\:pF$, $g = 0.8\:nS$, $E_L = -70\:mV$, $V_{reset} = -90\:mV$.\\

\noindent\textbf{Threshold Adaptation:} 
Equation~\ref{eqn:thesholdadapt} describes the adaptation of threshold 
voltage $V_{th}$~\cite{Fontaine_2014}.
\begin{equation}
    \label{eqn:thesholdadapt}
    \frac{dV_{th}}{dt} = -\frac{V_{th} - V_{th_0}}{\tau_0}
\end{equation}
where $V_{th}(t)$ is threshold voltage at any time $t$, $V_{th_0}$ is the
baseline threshold voltage, and $\tau_{0}$ is the threshold decay time constant.\\
Values used: $V_{th_0} = -55\:mV$, $\tau_{0} = 15\:ms$.

\subsection{Synapse Model}
\begin{figure}[!ht]
    \vspace{-8pt} 
    \centering
    \begin{minipage}[t]{0.4\textwidth}
        \centering
        \captionof{table}{Experimental data from RRAM~\cite{Lashkare_2020}.}
        \label{rram_expt_values}
        {\renewcommand{\arraystretch}{1.18}
        \begin{tabularx}{0.8\linewidth}{|c|Y|}
            \hline
            State & Conductance \(G\) ($\mu$S) \\ \hline
            1     & \(316.228\)         \\ \hline
            2     & \(199.526\)         \\ \hline
            3     & \(125.893\)         \\ \hline
            4     & \(63.096\)         \\ \hline
            5     & \(25.119\)         \\ \hline
            6     & \(12.589\)         \\ \hline
            7     & \(5.754\)        \\ \hline
            8     & \(3.981\)         \\ \hline
        \end{tabularx}
        }
    \end{minipage}
    \begin{minipage}[t]{0.58\textwidth}
        \vspace{0pt} 
        \centering
        \includegraphics[width=0.69\textwidth]{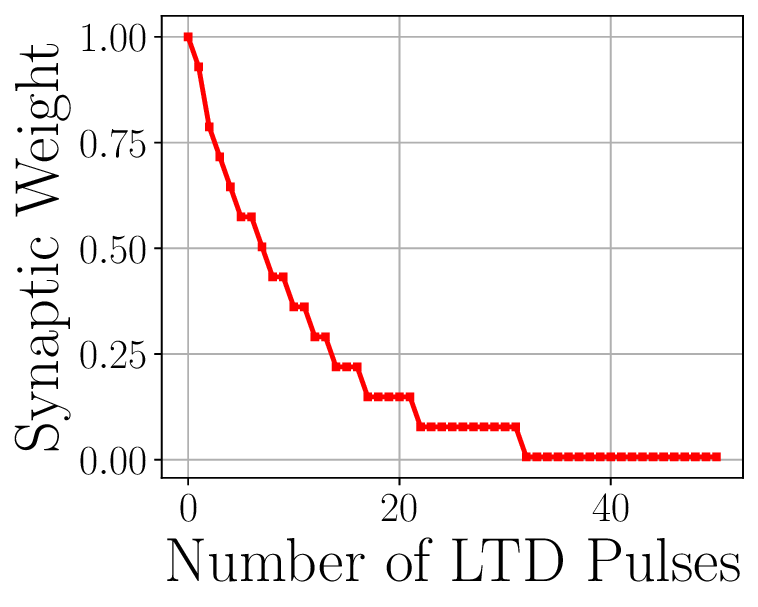}
        \caption{
        Non-linear synapse: weight vs.\ LTD pulse count. Initial weight is $1$ and number of states are $n=23$.}
        \label{fig:nonlinear_synapse}
    \end{minipage}
\end{figure}

Every input and output neuron is connected by a synapse (which is modelled by RRAM in hardware). It represents the 
strength of the connection between the two neurons. We normalise the synapse weight
(conductance of RRAM) so that it is between $10^{-3}$ (minimum synapse weight) and $1$ (maximum synapse weight). When we apply Long Term Potentiation (LTP) pulses, the synapse weight increases, and when we apply Long Term Depression (LTD) pulses, the synapse weight decreases.
\begin{figure}[!ht]
    \centering
    \begin{minipage}[b]{0.49\textwidth}
        \centering
        \includegraphics[width=0.7\textwidth]{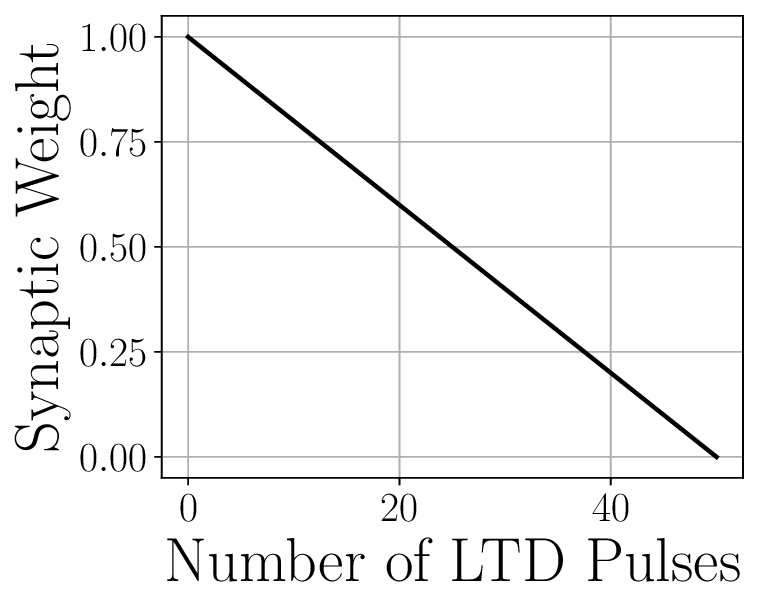}
        \caption{Ideal synapse: weight vs.\\LTD pulse count.}
        \label{ideal_synapse}
    \end{minipage}
    \begin{minipage}[b]{0.49\textwidth}
        \centering
        \includegraphics[width=0.7\textwidth]{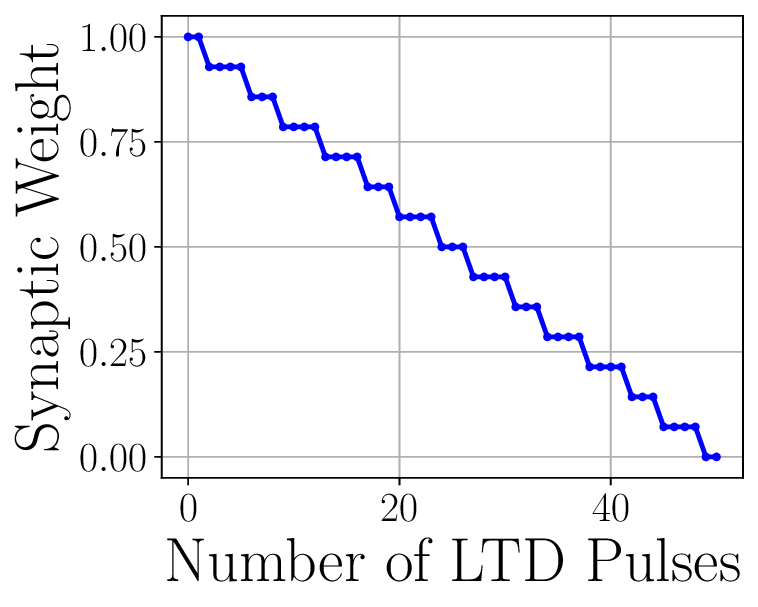}
        \caption{Linear synapse: weight vs.\\LTD pulse count.}
        \label{linear_synapse}
    \end{minipage}
\end{figure}
\begin{enumerate}
\item \textbf{Ideal Synapse:} This synapse has
an infinite number of states, and there is no non-linearity involved. Subsequent LTD (LTP) pulses uniformly
decrease (increase) the synapse weight (Fig.~\ref{ideal_synapse}, initial weight is set to $1$).
\item \textbf{Linear Synapse:} This is similar to an Ideal Synapse but is closer to 
a real RRAM. This synapse offers uniform weight change with LTD or LTP pulses, but there
are a finite number of states (Fig.~\ref{linear_synapse}, initial weight is set to $1$ and the number of states is $n=23$).
\item \textbf{Non-linear Synapse:} This synapse model accurately represents the 
real RRAM device (Pr$_{0.7}$Ca$_{0.3}$MnO$_3$ memristor), as subsequent pulses give a nonlinear weight change~\cite{Lashkare_2020}. We normalise the experimental RRAM conductance values (Table~\ref{rram_expt_values}), and 
fit a exponential function, described by equation~\ref{eqn:nonlinear}~\cite{Kim_2021}, which 
gives $w_i$, the weight of $i^{th}$ state for a $n$ state synapse.
\begin{equation}
    \label{eqn:nonlinear}
    w_i = w_{max} - \frac{w_{max} - w_{min}}{1-e^{-\nu}} 
    \left[ 1 - \exp\left( -\nu \left( 1 - \frac{i}{n} \right) \right) \right]
\end{equation}
$\nu$ is used to control the shape
of the curve, and $3.5 \le \nu \le 3.7$, models our experimental data most accurately.
Fig.~\ref{fig:nonlinear_synapse} shows the weight change of this synapse.
\end{enumerate}

\noindent Initially, all synapses are assigned the maximum weight of $1$. This allows the network to learn more easily, as all the connections are initially strong.
If the weights are assigned randomly, then certain neurons might never have a good connection to facilitate learning~\cite{biswas2016simpleefficientsnnperformance}.

\subsection{Learning Rule}

The learning rule specifies how much the weight should change between a pair
of pre- and post-synaptic spikes. We denote the time when a pre-synaptic
neuron fires by, $t_{pre}$ and when a post-synaptic neuron fires by,
$t_{post}$. We define $\Delta t$ as 
$t_{post} - t_{pre}$. The STDP function takes $\Delta t$ as input and returns the
weight change for the synapse between a corresponding pair of neurons. The learning rule 
is given by the equation~\ref{eqn:learning_rule}~\cite{Kasinski2006}.
\begin{equation}
    \label{eqn:learning_rule}
    \Delta w =
\begin{cases}
\eta \cdot F(\Delta t) \cdot (w - w_{\min})^{\gamma}, & \text{if } F(\Delta t) < 0  \\
\eta \cdot F(\Delta t) \cdot (w_{\max} - w)^{\gamma}, & \text{if } F(\Delta t) > 0 
\end{cases}
\end{equation}
\noindent Where, $\eta$ is the learning rate (we take $\eta \in [0.03, 0.13]$), 
$F(\Delta t) < 0$ indicates depression, $F(\Delta t) > 0$ 
indicates potentiation and $\gamma$ is a constant (we take $\gamma = 0.9$). 
The maximum synapse weight is $w_{\max} = 1$, and the minimum synapse weight is $w_{\min} = 10^{-3}$.
We describe four models of STDP ($F(\Delta t)$)~\cite{Abbott_2000,Bi10464,Wittenberg_2006}, as follows:
\begin{figure}[!ht]
    \centering
    \begin{minipage}[t]{0.3\textwidth}
        \vspace{0pt} 
        \centering
        \includegraphics[width=\textwidth]{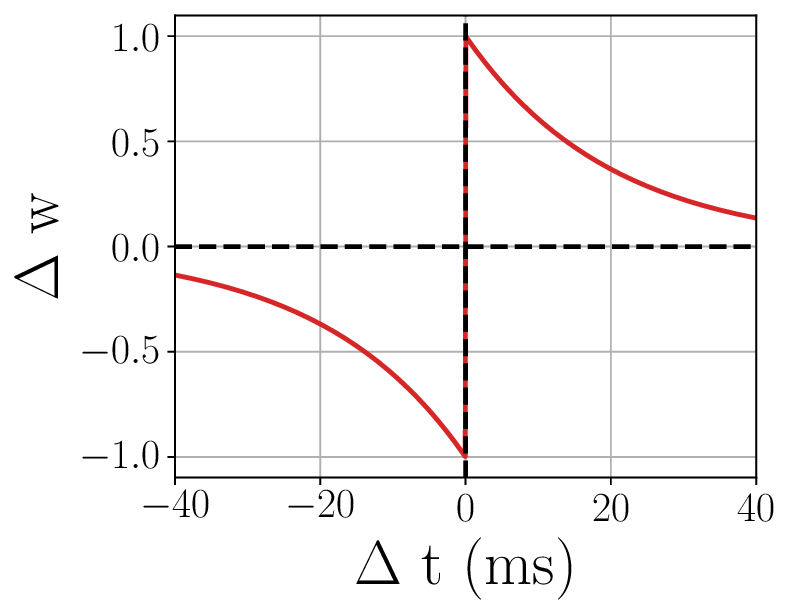}
        \caption{Conventional STDP curve (Eq.~\ref{eq:usual_stdp}).}
        \label{fig:usual_stdp}
    \end{minipage}
    \hspace{1em}
    \begin{minipage}[t]{0.3\textwidth}
        \vspace{0pt} 
        \centering
        \includegraphics[width=\textwidth]{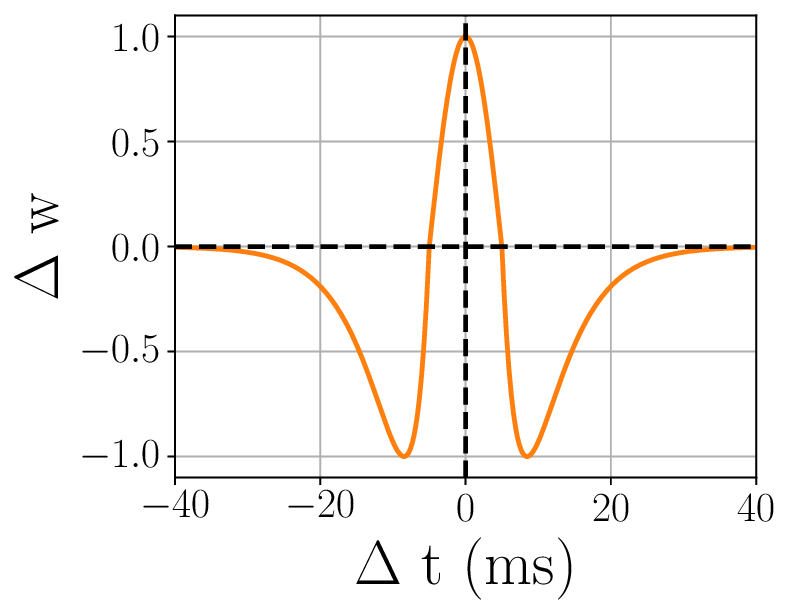}
        \caption{Cos STDP curve (Eq.~\ref{eq:cos_stdp}).}
        \label{fig:cos_stdp}
    \end{minipage}
    \hspace{1em}
    \begin{minipage}[t]{0.3\textwidth}
        \vspace{0pt} 
        \centering
        \includegraphics[width=\textwidth]{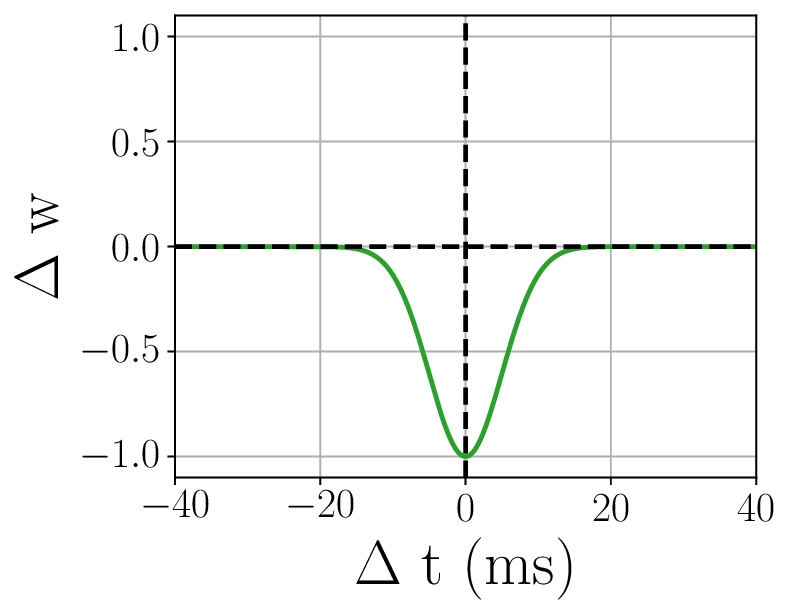}
        \caption{nGauss STDP curve (Eq.~\ref{eq:neg_gaussian_stdp}).}
        \label{fig:ngauss_stdp}
    \end{minipage}
    \vspace{-2pt} 
\end{figure}

\noindent\textbf{Conventional STDP: } 
\begin{equation}
    \label{eq:usual_stdp}
    F_{\text{conventional}}(\Delta t) =
\begin{cases}
A_{\text{up}} \cdot e^{-\frac{\Delta t}{\tau_{\text{up}}}}, & \text{if } \Delta t > 0 \ (\text{pre before post}) \\
A_{\text{down}} \cdot e^{\frac{\Delta t}{\tau_{\text{down}}}}, & \text{if } \Delta t < 0 \ (\text{post before pre})
\end{cases}
\end{equation}
\noindent Where $A_{up} = 0.8$, $A_{down} = -0.3$, $\tau_{up} = 5$, $\tau_{down} = 5$.
Fig.~\ref{fig:usual_stdp} shows the potentiation and depression curve for the Conventional STDP.
The lesser the time difference between post- and pre-neuron, the higher the weight increases. 
which indicates causation, the pre-neuron \textit{causes} the post-neuron to fire.
Similarly, in the other case, if a post-neuron fires before the pre-neuron, then 
the synapse weight decreases.\\

\noindent\textbf{Cos STDP: } 
\begin{equation}
    \label{eq:cos_stdp}
    F_{\text{cos}}(\Delta t) =
    \begin{cases}
        A_{\text{in}} \cos\!\left(\dfrac{\pi \Delta t}{2\tau_0}\right), & |\Delta t| \le \tau_0, \\[1ex]
        -A_{\text{out}} \Bigl(e^{-\alpha_1 (\Delta t-\tau_0)} - e^{-\alpha_2 (\Delta t-\tau_0)}\Bigr), & \text{Otherwise}, \\[1ex]
    \end{cases}
\end{equation}
\noindent Where     $\tau_0 = 1.5$, $A_{\text{in}} = 1$, $A_{\text{out}} = 4$,
$\alpha_1 = 0.2$, $\alpha_2 = 0.4$.
Fig.~\ref{fig:cos_stdp} shows the potentiation and depression curve for the Cos STDP,
It is symmetric, and if the time delay between the firing of two neurons is between $-\tau_0$
to $\tau_0$, the synapse weight increases, else it decreases.\\

\noindent\textbf{nGauss STDP: } 
\begin{equation}
    \label{eq:neg_gaussian_stdp}
    F_{\text{nGauss}}(\Delta t) = -A \exp\!\left(-\frac{\Delta t^2}{2\sigma^2}\right),
\end{equation}
\noindent Where $A = 1$, $\sigma = 5$.
Fig.~\ref{fig:ngauss_stdp} consists only of the depression region. There is no
potentiation. Hence, this rule can be used to unlearn the weights.\\

\begin{figure}[!ht]
    \vspace{-2pt} 
    \centering
\begin{minipage}[t]{0.3\textwidth}
        \centering
        \includegraphics[width=\textwidth]{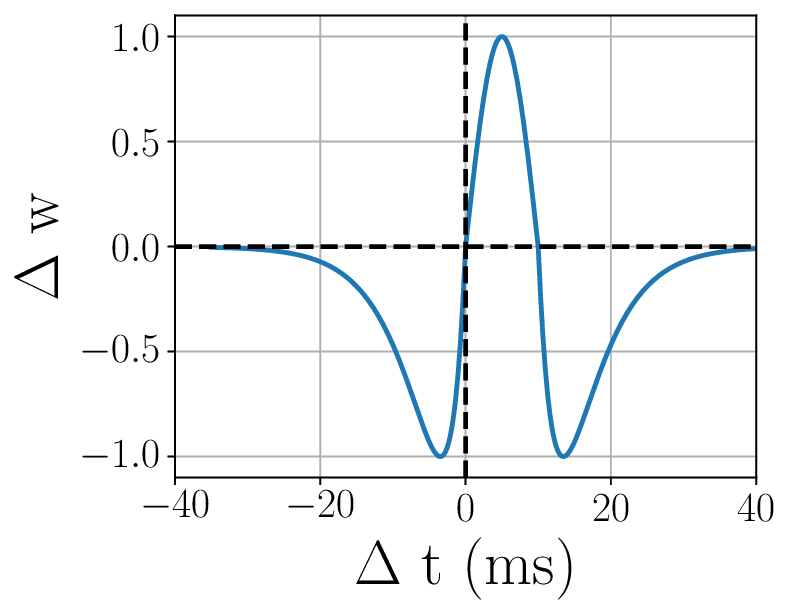}
        \caption{Sin STDP curve (Eq.~\ref{eq:sin_stdp}).}
        \label{fig:sin_stdp}
    \end{minipage}
    \hfill
    \begin{minipage}[t]{0.68\textwidth}
        \centering
        \includegraphics[width=\textwidth]{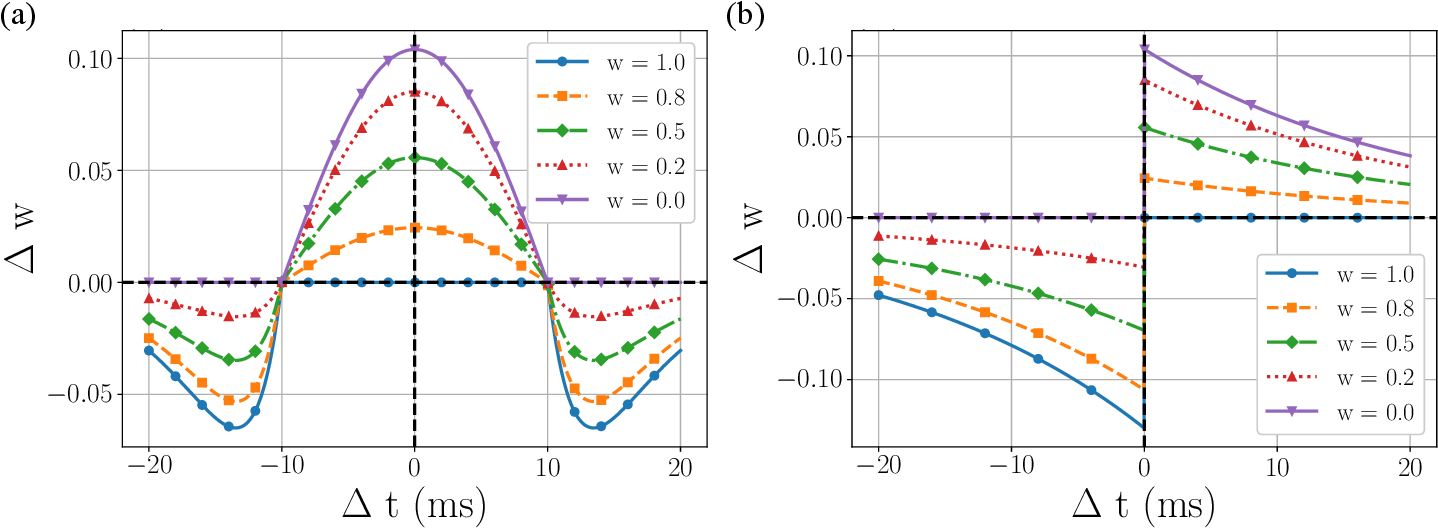}
    \caption{(a) LTD-LTP curve for Cos STDP, (b) LTD-LTP curve for Sin STDP.} 
    \label{fig:weightchange_cos_usual}
    \end{minipage}
    \vspace{-2pt} 
\end{figure}

\noindent\textbf{Sin STDP: } 
\begin{equation}
    \label{eq:sin_stdp}
    F_{\text{sin}}(\Delta t) =
    \begin{cases}
        -A_{\text{out}} \Bigl(e^{\alpha_1 \Delta t} - e^{\alpha_2 \Delta t}\Bigr), & \Delta t < 0, \\[1ex]
        A_{\text{in}} \sin\!\left(\dfrac{\pi \Delta t}{2\tau_0}\right), & 0 \le \Delta t \le 2\tau_0, \\[1ex]
        -A_{\text{out}} \Bigl(e^{-\alpha_1 (\Delta t-2\tau_0)} - e^{-\alpha_2 (\Delta t-2\tau_0)}\Bigr), & \Delta t > 2\tau_0,
    \end{cases}
\end{equation}
\noindent Where $\tau_0 = 5$, $A_{\text{in}} = 1$, $A_{\text{out}} = 4$,
$\alpha_1 = 0.2$, $\alpha_2 = 0.4$.
Fig.~\ref{fig:sin_stdp} shows the potentiation and depression curve for the Sin STDP,
If the pre-neuron fires after the post-neuron, or the post-neuron fires more than $2\tau_0$ after the pre-neuron, then the synapse weight decreases. Otherwise, it increases.

\noindent Fig.~\ref{fig:weightchange_cos_usual} shows the comparison between the weight change of a synapse that follows the learning rule given by equation~\ref{eqn:learning_rule}, and the two STDPs, Cos STDP (equation~\ref{eq:cos_stdp}) and Conventional STDP (equation~\ref{eq:usual_stdp}).
We assume five initial weights of the synapse and see how the weights change.
In both the figures, Fig.~\ref{fig:weightchange_cos_usual}(a) and Fig.~\ref{fig:weightchange_cos_usual}(b), the maximum weight synapse, can only decrease and the minimum weight synapse can only increase.

\subsection{MNIST Dataset and Input Encoding}
The MNIST dataset consists of images of handwritten digits~\cite{726791} used to evaluate image classification models. Each image is a 28 \(\times\) 28 grayscale matrix with pixel intensities from 0 (black) to 255 (white). We flatten each image to a 784-vector and normalize it to the range \([0, 1]\). So, each entry $p \in [0,1]$ is then converted to a firing frequency \(f\) by:
\begin{equation}
    f = p \times (f_{\text{max}} - f_{\text{min}}) + f_{\text{min}},\quad f_{\max}=70,\;f_{\min}=5
    \label{eq:mnist_eqn}
\end{equation}
Brighter pixels result in higher spiking frequencies. The encoded frequencies generate spike trains over a fixed duration, \textit{training\_duration}, which is set to 100 by default. So this produces a binary spike-train matrix 
$
  M\in\{0,1\}^{784\times100},
$
where \(M_{ij}=1\) if \(i^{th}\) neuron fired at the \(j^{th}\) millisecond, and \(M[i][j] = 0\) otherwise.

\subsection{Evaluation}
For five-class classification, our two-layer SNN ($784 \text{ input} \rightarrow 80 \text{ output}$ neurons) achieves an accuracy of $92.7\%$ when trained on 100 images, and tested on 1500 unseen images. As illustrative baselines, we train the following models on the same data:

\begin{enumerate}
    \item \textbf{ANN:} Input layer (784 units) $\rightarrow$ Dense(128 units, ReLU activation) $\rightarrow$ Dense(5 units, softmax activation). 
    
    \item \textbf{CNN:}  
        Input 28$\times$28$\times$1 $\rightarrow$ Conv2D(32 filters, 3$\times$3 kernel window, ReLU activation) $\rightarrow$ MaxPooling2D(2$\times$2) $\rightarrow$ Conv2D(64 filters, 3$\times$3 kernel window, ReLU activation) $\rightarrow$ MaxPooling2D(2$\times$2) $\rightarrow$ Flatten $\rightarrow$ Dense(128 units, ReLU activation) $\rightarrow$ Dense(5 units, softmax activation).  
    \end{enumerate}
\noindent For both models, training is done using the Adam optimizer, with a 0.001 learning rate, a categorical cross-entropy loss function, a 64 batch size, and 10 epochs.
\section{Results}

\subsection{Changing STDP Rule }
We evaluate our SNN on MNIST with ideal synapses using Conventional, Cos, Sin, and nGauss STDP across five-, seven-, and ten-class tasks. The best-performing rule is selected based on accuracy. For five classes we use 100 training images (200 for seven and ten). As nGauss alone always reduces synapse weights, we apply it briefly (with small parameters) to a random 10\% of training images each epoch to promote unlearning, then switch back to the primary rule. Initial weights are 1 for Conventional and Sin STDP, but random for Cos STDP to allow any weight change otherwise all-1 weights would stay fixed unless spike time differences are huge.

\begin{figure}[!ht]
  \centering
  \begin{minipage}[t]{0.46\textwidth}
    \vspace{0pt}
    \centering
    \includegraphics[width=\textwidth]{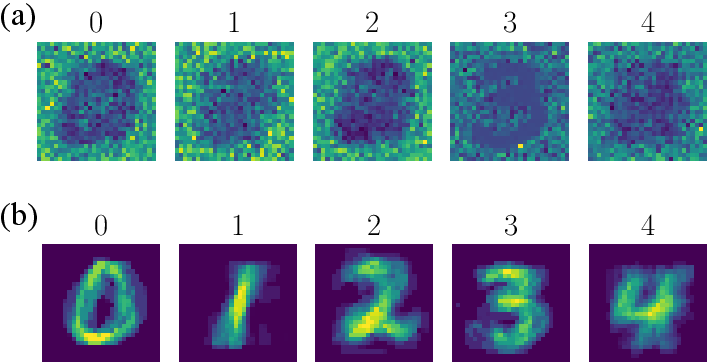}
    \caption{Synapse weights visualised for (a) Cos STDP and (b) Sin STDP} 
    \label{fig:cos_stdp_weights_final}
  \end{minipage}
  \hfill
  \begin{minipage}[t]{0.53\textwidth}
    \vspace{0pt}
    \centering
    \captionof{table}{Accuracies (\%) observed in the classification of 1500 test images.}
    \label{stdp_rule_change}
    {\renewcommand{\arraystretch}{1.35}
    \begin{tabular}{|c|c|c|c|}
    \hline
    Classes & \shortstack{Conventional\\STDP (\%)} & \shortstack{Cos\\STDP (\%)} & \shortstack{Sin\\STDP (\%)} \\ \hline
    5       & 92.73                                & 25.72                       & 86.40                       \\ \hline
    7       & 87.32                                & 17.60                       & 80.44                       \\ \hline
    10      & 76.27                                & 13.45                       & 69.86                       \\ \hline
    \end{tabular}
    }
\end{minipage}

  \vspace{-4pt} 
\end{figure}

Table~\ref{stdp_rule_change} shows the accuracies achieved using the respective STDP rules. So, Cos STDP can be ruled out, as it performs poorly. The visualization of synapse weights for Cos STDP (Fig.~\ref{fig:cos_stdp_weights_final}(a)) shows a ghosting effect of expected digits, indicating unlearning. Since for Cos STDP, the initial synapse weights are assigned randomly, for \(\Delta t > 0\), most causal neuron synapse weights increase. However, when \(\Delta t < 0\) there is a high chance that the synapse weight increases again, leading to higher weights for synapses of neurons at the edges (Fig.~\ref{fig:cos_stdp_weights_final}(a)).

\noindent Sin STDP performs better than Cos but not as well as Conventional (Fig.~\ref{fig:cos_stdp_weights_final}(b) and Fig.~\ref{learned_weights_heatmap_overall_workflow}(a)). It suggests that a larger weight for neurons with \(\Delta t > 0\) leads to better accuracy. It should be noted that if \(\tau_0\) is smaller, then more instances occur where synapse weights decrease for \(\Delta t > 0\), so the accuracy drops.

\subsection{Changing Synapse Model and Trade-offs}
\begin{figure}[!ht]
    \centering
    \includegraphics[width=\textwidth]{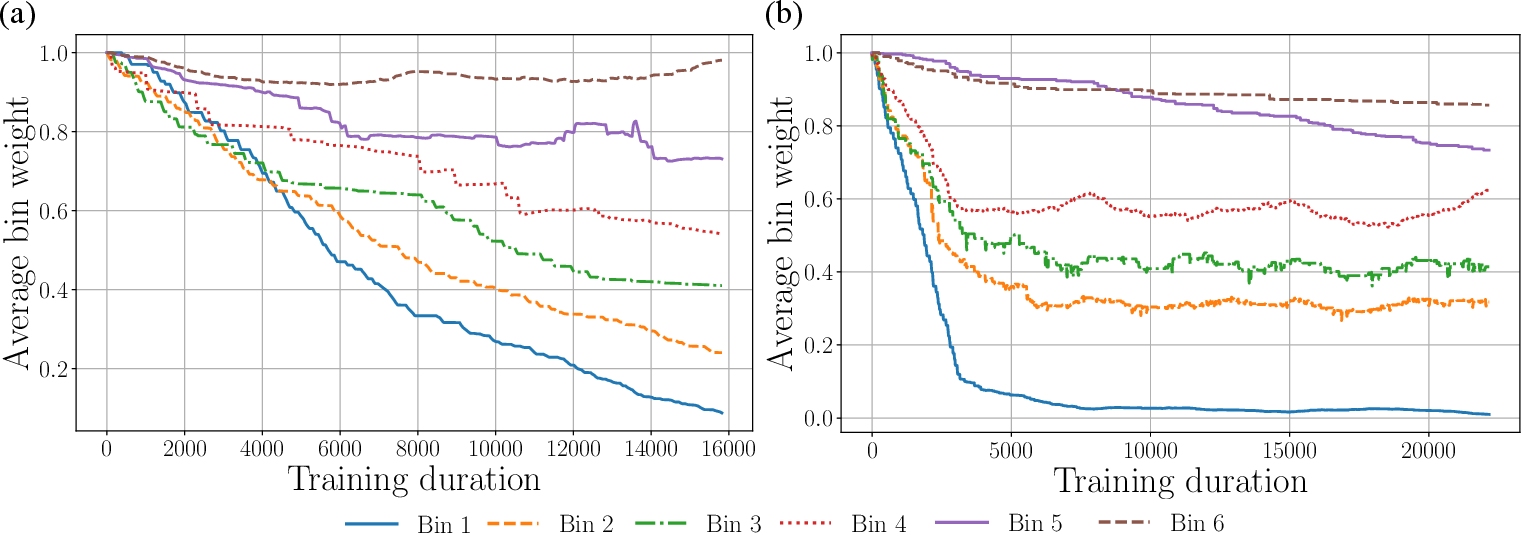}
    \caption{Weight evolution for (a) an Ideal Synapse and (b) Non-linear Synapse based SNN for classifying digits 0 to 4.} 
    \label{fig:weight_evol_ideal_nl}
\end{figure}

\begin{figure}[!ht]
    \centering
    \includegraphics[width=\textwidth]{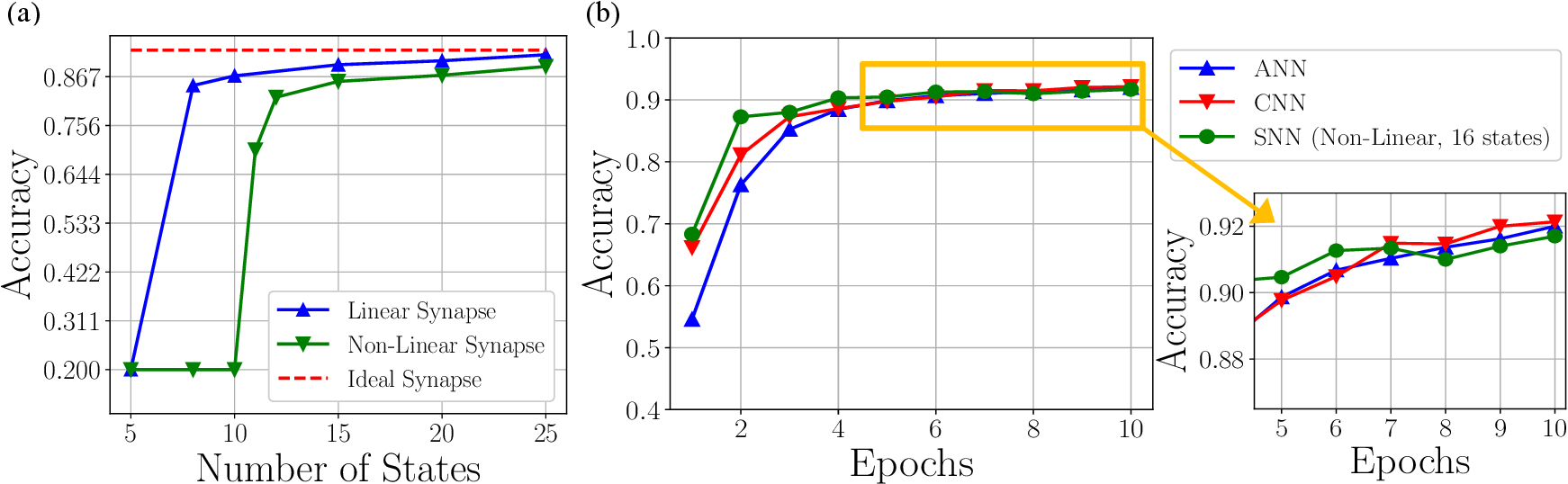}
    \caption{(a) Accuracy vs. Number of States, for five-class MNIST, (b) Comparison of SNN with ANN and CNN.} 
    \label{linear_nonl_acc_state_snnannaccuracy}
\end{figure}

The best STDP rule comes out to be \textbf{Conventional STDP}, so we proceed with it and evaluate on the synapse models. Table~\ref{mnist_eval_table} gives the architecture and average accuracy obtained by classifying on the MNIST dataset. 
\begin{table}[!ht]
    \vspace{-8pt} 
    \centering
    \caption{MNIST classification using Conventional STDP for various synapses}
    \label{mnist_eval_table}
    \resizebox{\textwidth}{!}{
    \begin{tabular}{|c|cc|cc|cc|}
    \hline
    \multirow{2}{*}{Number of Classes} &
      \multicolumn{2}{c|}{Ideal Synapse} &
      \multicolumn{2}{c|}{Linear Synapse} &
      \multicolumn{2}{c|}{Non-linear Synapse} \\ \cline{2-7} 
     &
      \multicolumn{1}{c|}{Output Neurons} &
      Accuracy (\%) &
      \multicolumn{1}{c|}{Output Neurons} &
      Accuracy (\%) &
      \multicolumn{1}{c|}{Output Neurons} &
      Accuracy (\%) \\ \hline
    5  & \multicolumn{1}{c|}{80}  & 92.73 & \multicolumn{1}{c|}{80}  & 91.67 & \multicolumn{1}{c|}{60}  & 91.07 \\ \hline
    6  & \multicolumn{1}{c|}{160} & 88.93 & \multicolumn{1}{c|}{140} & 87.88 & \multicolumn{1}{c|}{140} & 84.73 \\ \hline
    7  & \multicolumn{1}{c|}{160} & 87.32 & \multicolumn{1}{c|}{140} & 85.58 & \multicolumn{1}{c|}{140} & 83.98 \\ \hline
    8  & \multicolumn{1}{c|}{160} & 87.03 & \multicolumn{1}{c|}{140} & 86.58 & \multicolumn{1}{c|}{140} & 82.40 \\ \hline
    10 & \multicolumn{1}{c|}{160} & 76.27 & \multicolumn{1}{c|}{160} & 74.46 & \multicolumn{1}{c|}{120} & 71.20 \\ \hline
    \end{tabular}
    }
    \vspace{-6pt} 
\end{table}
The number of input neurons is 784, and the number of test images for computing accuracy is 1500. The learning rate varies between $0.03$ to $0.13$, as we go from Ideal Synapse to a Non-linear Synapse (with 25 states). It can be clearly seen in all cases, the accuracy drops from moving from Ideal to Linear to Non-linear Synapse.
The weight evolution for five-class classification can be seen in Fig.~\ref{fig:weight_evol_ideal_nl}, the final synapse weights are grouped 
together into six bins, based on their final weights, and for two epochs, the average weight of each bin is recorded.
We see that for the Ideal Synapse (Fig.~\ref{fig:weight_evol_ideal_nl}(a)), the weight decrease for each bin is uniform, whereas for the Non-linear Synapse (with 25 states), the synapse reach their final weight in less time (Fig.~\ref{fig:weight_evol_ideal_nl}(b)). Both in the case of Linear and Non-linear Synapse, varying the number of states from $25$ to $5$ decreases the accuracy, for a particular classification task,
as seen in Fig.~\ref{linear_nonl_acc_state_snnannaccuracy}(a). However, with an increase in the number of states, the Non-linear Synapse reaches the accuracy
levels of the Linear Synapse.

\noindent The measured time of training the model on AMD Ryzen 7840HS on five-class MNIST with Conventional STDP and Non-linear Synapse for one epoch, is 3 min $\pm$ 5s (averaged over 10 epochs) and the inference of a single sample, comes out to be 6.553 $\pm$ 0.0058 ms (averaged over 7500 inferences).

\subsection{Comparison with ANN and CNN}
We train an ANN and a CNN to classify digits from 0 to 4, and report the accuracy as epochs increase.
We compare these two with an SNN with Non-linear Synapse with 16 states respectively (Fig.~\ref{linear_nonl_acc_state_snnannaccuracy}(b)). Note that we keep the training and testing images the same for all models. So, both CNN and ANN are expected to perform poorly as there are only 100 training images and 1500 testing images. However, with more training images, CNN can achieve an accuracy of 99\%~\cite{726791}, but the network architecture will not be simple. All accuracy numbers reported throughout the paper were averaged over 5 independent runs (with standard deviation < 1\%).

\section{Discussion}
\begin{figure}[!ht]
    \centering
    \includegraphics[width=0.88\textwidth]{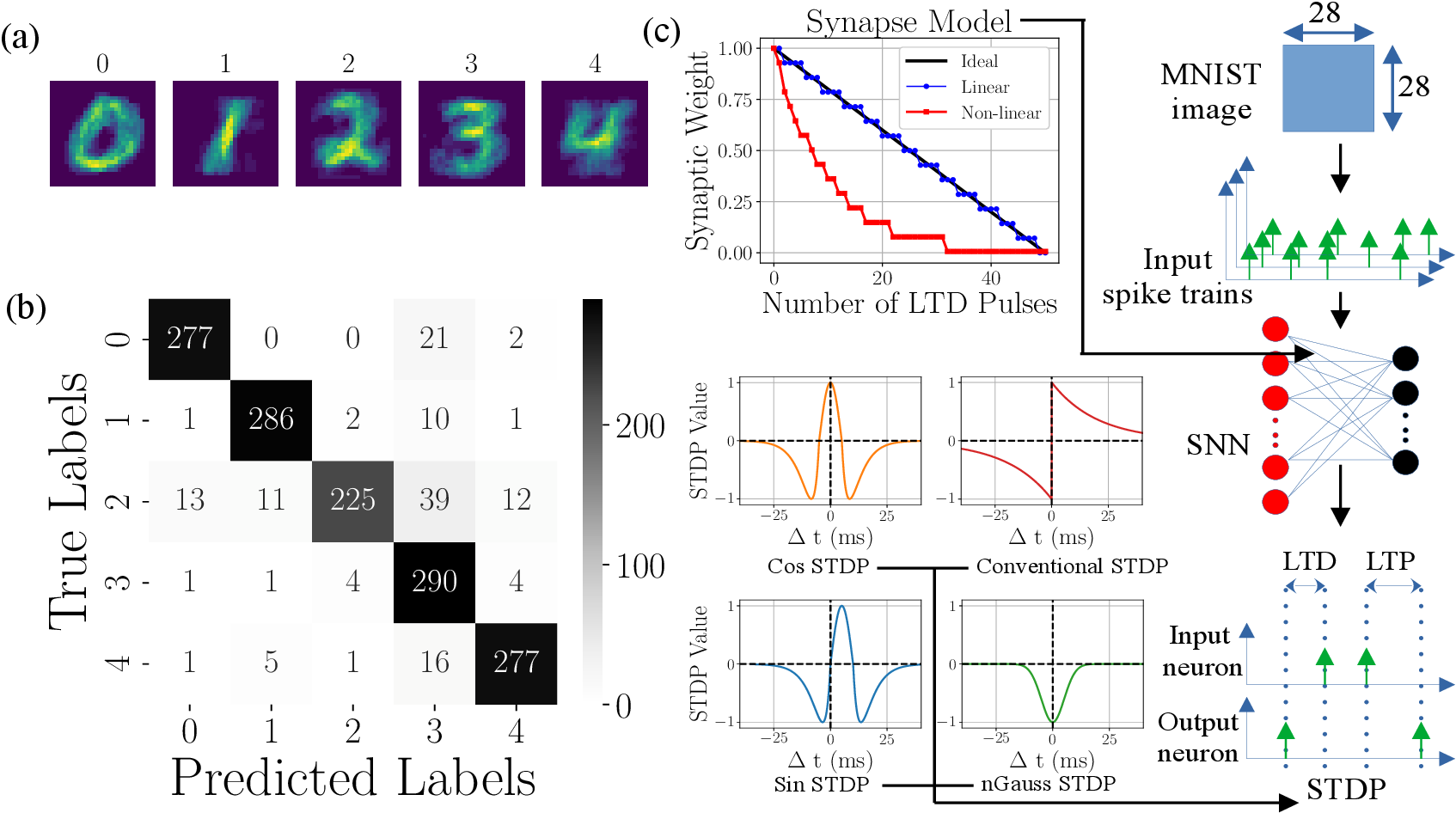}
    \caption{(a) Weights learned by our network, (b) Confusion matrix for our SNN on five-class MNIST, (c) Overall workflow of our methodology.}
    \label{learned_weights_heatmap_overall_workflow}
    \vspace{-5pt}
\end{figure}

SNN reaches a good accuracy in a small number of epochs, with a small amount of training data, and with a simple architecture (Fig.~\ref{learned_weights_heatmap_overall_workflow}(c) for overview of the methodology). From Fig.~\ref{learned_weights_heatmap_overall_workflow}(b) it is seen that for five-class MNIST, our model usually confuses two with three, which produce very similar spike-train pattern under our rate encoding. Fig.~\ref{learned_weights_heatmap_overall_workflow}(a) shows that the learned weights for classes 2 and 3 significantly overlap, which explains the high misclassification between these two classes.
\noindent Unlike conventional ANNs/CNNs that process dense, synchronous activations, our STDP based SNN leverages sparse, event-driven spikes~\cite{fncom.2015.00099}. This allows it to converge in fewer epochs on small datasets and, when mapped to memristive crossbars, consume less energy per image than a GPU-trained CNN~\cite{50211040}. In practice, SNNs excel on always-on, low-power platforms, for example, edge Internet of Things devices with Dynamic Vision Sensor (DVS) cameras, real-time sensory processing in robotics, or ultra-efficient neuromorphic chips where conventional nets (with their high multiply-accumulate counts ) are impractical or wasteful~\cite{50211040,10890793}. Although we focused on the canonical MNIST benchmark, future work will extend our STDP-synapse framework to harder datasets (Fashion-MNIST, CIFAR-10) and to temporal DVS streams and the deployment will be studied on hardware memristor crossbar array to get an actual measurement of power/memory used.
\section{Conclusion}
Our study shows that neuromorphic computing with SNNs is effective for pattern recognition. Among the four STDP variants examined, Conventional STDP performed best on five-, seven-, and ten-class MNIST classification, achieving over 90\% accuracy in five-class MNIST classification. This shows the importance of causal learning, where synaptic weights increase when pre-synaptic neurons trigger post-synaptic firing. We observed trade-offs between classification performance and hardware realism of synapses (Linear $\rightarrow$ Non-linear). The Non-linear and Linear models achieved 91.07\% and 91.67\% accuracy respectively, while the Ideal synapse model reached 92.73\%. Accuracy dropped significantly when synapse states were reduced from 25 to 5, indicating that hardware with few memory states performs poorly. Despite these limitations, our SNN architecture offers several advantages over traditional neural networks. It converges in fewer training epochs and achieves decent accuracy with only 100 training images for five-class classification. This efficiency makes SNNs attractive for applications like autonomous systems and edge devices, where conserving power is essential.

\begin{credits}
\subsubsection{\ackname} 
This work is supported in part by Anusandhan National Research Foundation (ANRF) Inclusivity Research Grant (IRG) ANRF\slash IRG\slash 2024\slash 000139\slash ENS and startup research grant by IIT Gandhinagar. This work is supported from the tools received under the C2S project from MEITY, Government of India.
\subsubsection{\discintname}
We have no competing interests to declare that are relevant to the content of this article.
\end{credits}
\bibliographystyle{splncs04}  
\bibliography{references} 
\end{document}